\documentclass[10pt,twocolumn,letterpaper]{article}

\usepackage[pagenumbers]{cvpr} 

\usepackage{xfrac}
\usepackage{soul}
\newcommand{\hlc}[2][yellow]{{%
    \colorlet{foo}{#1}%
    \sethlcolor{foo}\hl{#2}}%
}

\usepackage[dvipsnames]{xcolor}

\definecolor{cvprblue}{rgb}{0.21,0.49,0.74}
\usepackage[pagebackref,breaklinks,colorlinks,allcolors=cvprblue]{hyperref}

\usepackage{amsmath}
\usepackage{algorithm}
\usepackage{algpseudocode}

\usepackage{graphicx}
\usepackage{multirow}
\usepackage{colortbl} 
\usepackage{booktabs} 

\usepackage{arydshln} 
\usepackage{tcolorbox}

\definecolor{isabelline}{rgb}{0.96, 0.96, 0.86}
\definecolor{lightblue}{rgb}{0.68, 0.85, 0.9}

\usepackage{pifont}
\newcommand{\cmark}{\ding{51}}%
\newcommand{\xmark}{\ding{55}}%

\newcommand{\vlms}{VLMs}

\newcommand{\oursfull}{\textbf{Vi}deo \textbf{T}emporal \textbf{E}vidence \textbf{D}istillation for Video Understanding}
\newcommand{\ours}{\textsc{ViTED}}

\newcommand{\ourmodel}{\textsc{ViTED}}

\def\paperID{xxxxx} 
\def\confName{CVPR}
\def\confYear{2025}

\title{\ours~: Video Temporal Evidence Distillation}

\author{%
Yujie Lu$^{1,2}$ \hspace{3mm} Yale Song$^{2}$ \hspace{3mm}William Wang$^{1}$ \hspace{3mm} Lorenzo Torresani$^{2}$ \hspace{3mm} Tushar Nagarajan$^{2}$\\
$^1$UC Santa Barbara \hspace{3mm} $^2$FAIR, Meta \hspace{3mm}
}

\begin{document}
\maketitle
\begin{abstract}
We investigate complex video question answering via \emph{chain-of-evidence} reasoning --- identifying sequences of temporal spans from multiple relevant parts of the video, together with visual evidence within them.
Existing models struggle with multi-step reasoning as they uniformly sample a fixed number of frames, which can miss critical evidence distributed nonuniformly throughout the video. Moreover, they lack the ability to temporally localize such evidence in the broader context of the full video, which is required for answering complex questions. We propose a framework to enhance existing VideoQA datasets with evidence reasoning chains, automatically constructed by searching for optimal intervals of interest in the video with supporting evidence, that maximizes the likelihood of answering a given question.
We train our model (\ourmodel) to generate these evidence chains directly, enabling it to both localize evidence windows as well as perform multi-step reasoning across them in long-form video content.
We show the value of our \emph{evidence-distilled} models on a suite of long video QA benchmarks where we outperform state-of-the-art approaches that lack evidence reasoning capabilities.
\end{abstract}
    
\section{Introduction}
\label{sec:intro}

Video Question Answering (VideoQA) is a critical task towards general-purpose video understanding, for which large vision-language models (VLMs)~\cite{zhang2023videollamainstructiontunedaudiovisuallanguage, wang2024internvideo2, li2024llavaonevision}  have recently demonstrated strong performance on a variety of benchmarks~\cite{xiao2021next,xiao2024itrustanswervisually,wu2021star_situated_reasoning,pătrăucean2023perceptiontest,li2024mvbench,rawal2024cinepile}. 
Despite their popularity, these models typically excel at questions for which information is readily available throughout the video (e.g., high-level actions, colors/counts of objects, etc.), but struggle on long-form videos~\cite{pătrăucean2023perceptiontest, mangalam2023egoschema, he2024mmworld, rawal2024cinepile} and on questions that require gathering and aggregating evidence from the full video~\cite{xiao2024itrustanswervisually}.
For instance, answering a question like, ``Why does the baby put their hand in their mouth at the beginning of the video?" requires more than just locating the action --- it involves an intricate approach that gathers contextual clues both before and after the action to infer the underlying reason. For instance, examining what occurred just before or after the baby put their hand in their mouth and observing how they performed the action (e.g., slowly or quickly) can offer valuable insights into their intent.

\begin{figure}[t]
\centering
\includegraphics[width=\columnwidth]{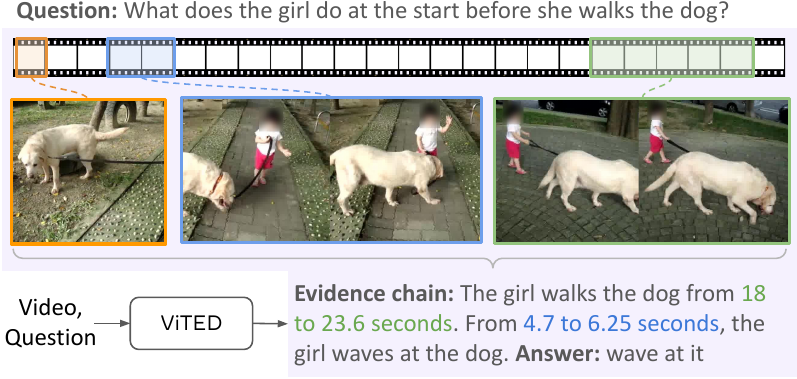}
\caption{
\textbf{Main idea.} We produce multiple, temporally localized pieces of evidence (the ``evidence chain’’) to support complex reasoning in VideoQA. Our \ours~model is trained to generate this evidence chain to enable temporally-grounded chain-of-thought reasoning in video.
}
\label{fig:teaser}
\vspace{-4mm}
\end{figure}

Current models face limitations in effectively processing the temporal relationships among visual frames, as well as in bridging the gap between a question and the evidence needed to answer it.
On the one hand, current models do not temporally ground their responses, and rather sample a fixed number of video frames at regular intervals, potentially
missing important moments from the video (e.g., failing to identify the short window when the girl waves at dog in Fig.~\ref{fig:teaser}, left). Consequently, these models struggle to answer questions requiring complex \textit{temporal understanding}, e.g., ``what did the girl do right before interacting with the dog?" in a video where she approaches, picks up a ball, then engages with the dog; and \textit{multi-hop reasoning}, e.g., ``why did the baby put his hand in his mouth?" in a video where the mother feeds him with a spoon, followed by his uncomfortable expression and an attempt to remove the food.

On the other hand, \emph{temporal grounding} models can indeed localize text queries to specific intervals in a video~\cite{ren2024timechat,huang2024litalanguageinstructedtemporallocalization,lin2023univtgunifiedvideolanguagetemporal}, but they require the text in advance --- i.e., they can localize the evidence needed to answer the question if already provided with the evidence text, but they cannot identify and ground this evidence based on the question alone.
For example, it might identify the moment when the dog approaches the girl (Fig.~\ref{fig:teaser}), but it may not establish how this action relates to the question (``why did she wave?'').

To address these limitations, we propose to consolidate evidence generation, temporal grounding and question answering into one model that we call \oursfull~(\ours). Our model is a temporally-aware VLM that is trained to generate both answers to given questions, as well as temporally grounded \emph{evidence chains} to support the answer --- time intervals of the video together with textual clues within the temporal span (see Fig.~\ref{fig:teaser}, bottom).
Since this kind of evidence data is not readily available, we present a framework to automatically synthesize high-quality evidence chains on top of existing VideoQA datasets.
Specifically, we generate a pool of evidence containing textual evidence relevant to the question, extracted from video segments of various lengths and at multiple levels of detail (e.g., short clips with single actions, to high level activities across the full video).
We then present a search-and-refinement algorithm over this evidence pool to find optimal sequences of evidence that are the most predictive of the correct answer.
Finally, we augment the original VideoQA training data with our generated evidence chains, and train our model to predict both the answer and the evidence chain that supports it, thereby distilling the ability to localize and generate temporal evidence into the VLM.

We demonstrate the effectiveness of \ours~ on 6 representative VideoQA benchmarks, CinePile~\cite{rawal2024cinepile}, PerceptionTest~\cite{pătrăucean2023perceptiontest}, NExT-QA~\cite{xiao2021next}, STAR~\cite{wu2021star_situated_reasoning}, MVBench~\cite{li2024mvbench}. We show that \ours~ is on par with or outperforms state-of-the-art models trained with 10$\times$ more video instruction data. Additionally, we show that \ours~ provides the most faithful and interpretable temporal evidence chain for the answer compared with existing VideoLLMs through human studies. Finally, \ours~ achieves SOTA zero-shot performance on NExT-GQA~\cite{xiao2024itrustanswervisually} --- a benchmark squarely focused on temporally grounded VideoQA --- surpassing GPT-4 driven agent approaches, highlighting our generalizable evidence grounding capability.

In summary, we propose an approach that integrates evidence generation, grounding and reasoning towards complex video understanding.
Our main contributions are:
\begin{itemize}
    \item We propose a novel framework to generate and search for evidence chain-of-thought data from existing VideoQA datasets.
    \item We propose an \emph{evidence distillation} approach to train a temporally-aware video model on our high-quality evidence data.
    \item Our \ours~ sets new SOTA results among models of the same size on four VideoQA benchmarks, and surpasses GPT-4 driven agent on NExT-GQA, while providing high-quality explanations for its predictions.
\end{itemize}

\begin{figure*}[!t]
\centering
\includegraphics[width=\textwidth]{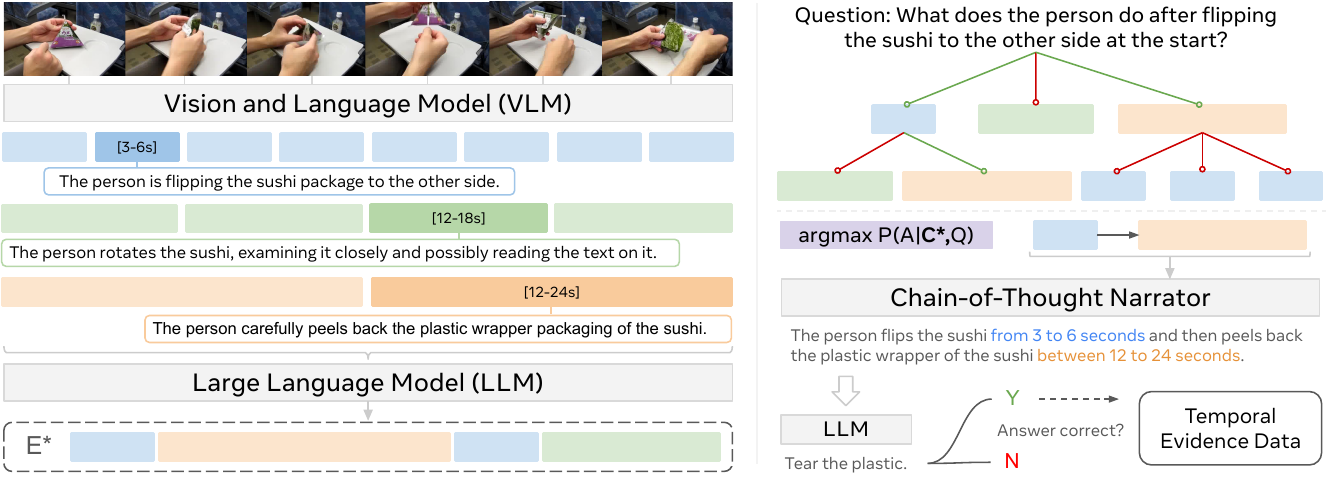}

\caption{\textbf{Overview of \ours~
evidence generation framework.}
There are three main stages: (1) We first generate the evidence pool --- detailed captions for segments at multiple granularities --- and rank them based on relevance to the question (left, Sec.\ref{sec:evidence_pool}) (2) Next, we search over the evidence pool to derive evidence chains that are most predictive of the target answer, and summarize it into a coherent and logical chain-of-thought (top-right, Sec.~\ref{sec:evidence_search}) (3) Finally, if the evidence chain successfully leads to the correct answer, we add it to our dataset for training our model (bottom-right, Section~\ref{sec:evidence_distill})
}

\label{fig:evidence_overview}
\vspace{-3mm}
\end{figure*}
\section{Related Work}
\label{sec:related}
\subsection{Video Understanding with LLMs}
Recent LLM-based video models~\cite{zhang2023videollamainstructiontunedaudiovisuallanguage, maaz2024videochatgptdetailedvideounderstanding, lin2024vilapretrainingvisuallanguage,li2024llavaonevision,wang2024internvideo2,zhang2024llavavideo} excel in video QA, but they are not temporally sensitive --- they process uniformly sampled video frames, potentially skipping key moments, and are unable to perform multi-step reasoning.
Recent efforts~\cite{ren2024timechat, huang2024litalanguageinstructedtemporallocalization,wang2024hawkeyetrainingvideotextllms} incorporate explicit time-aware representations for video-text grounding, however they stop short of evaluating models on general VideoQA.
In contrast, we consolidate evidence generation, grounding and question answering into a single model.

Agent-based or tool-assisted VLMs gather relevant video tokens for a given question, e.g., retrieval-augmented~\cite{wang2024videoagent,Shang2024TraveLERAM}, memory-augmented~\cite{he2024MA-LMM,song2024moviechat,fan2024videoagentmemory}, and modular reasoning agents~\cite{min2024morevqaexploringmodularreasoning,yu2023sevila}.
In contrast, ours focuses on generating and localizing relevant evidence to support answering a question. In particular, we distill evidence collection and reasoning capabilities into a single-pass model, bypassing the need for multiple independent modules or API calls, while still improving performance.

\subsection{Chain-of-Thought Reasoning in Videos}
Chain-of-thought (CoT)~\cite{wei2023chainofthoughtpromptingelicitsreasoning} has been widely used to enhance the multi-step reasoning ability of LLMs. Various strategies have been proposed to ensure valid and logical reasoning paths through problem decomposition~\cite{Zhou2022LeasttoMostPE}, deliberate search~\cite{shunyu2024treeofthought}, and majority voting~\cite{Wang2022RationaleAugmentedEI,wang2023selfconsistencyimproveschainthought}. Some works use knowledge distillation to enhance smaller models with the reasoning ability of larger models~\cite{magister-etal-2023-teaching,li2024symbolicchainofthoughtdistillationsmall,shridhar-etal-2023-distilling}, or internalize this reasoning through implicit distillation~\cite{deng2024explicitcotimplicitcot}.

In VLMs, CoT-based techniques~\cite{lu2022learn,zhang2024multimodalchainofthoughtreasoninglanguage} improve reasoning by generating textual rationales or synthesizing multimodal infillings~\cite{rose2024visualchainthoughtbridging}. However, CoT for video understanding is under-explored. VIP~\cite{himakunthala2023letsthinkframeframe} enhances CoT in LLMs and VLMs for video prediction, while methods like VSOR-CoT~\cite{tang2024cardiffvideosalientobject} improve video saliency prediction by reasoning about salient objects. MotionEpic~\cite{Fei2024VideoofThoughtSV} decomposes video tasks for better question answering using a Video-of-Thought framework. To the best of our knowledge, we are the first to \textit{distill} chain-of-thought capabilities in \vlms, specifically for video understanding.

\subsection{Visual Evidence}
In image understanding, prior work has explored gathering visual evidence in the form of region of interest~\cite{you2023ferret}, interface layout~\cite{qian2024vui} and programs~\cite{hu2024visualprogramdistillationdistilling}.
In videos, prior work largely focuses on \emph{frame-sampling} that varies sampling rate of the video depending on the content~\cite{TanKoala2024}, employs key-frame extraction pipelines either through off-the-shelf approaches~\cite{Katna,maaz2024videochatgptdetailedvideounderstanding} or based on learned text-frame similarity~\cite{liang2024keyvideollmlargescalevideokeyframe,yu2023sevila}. Approaches for highlight detection cast visual evidence as a single time-window that matches a text description~\cite{ren2024timechat,song2024moviechat,moon2023querydependentvideorepresentationmoment}.
In contrast to the above, we propose to treat visual evidence in videos differently, as a series of temporally grounded descriptions that chain together to entail the answer to a question.
\section{Approach}
\label{sec:sec3_approach}
Our goal is to enable evidence-based video reasoning in VLMs by generating and distilling evidence chain data (i.e., time intervals with textual clues that support question-answering) into a model. This approach is crucial for datasets with complex reasoning questions. For example, in a study on NExT-QA, we found $54\%$ of questions required localizing and reasoning over one or more salient windows (see Appendix~\ref{app:preliminary_study}). In short, we convert the standard question answering process (Q$\rightarrow$A) into an evidence-based reasoning one (Q$\rightarrow$Evidence + A).

Existing VideoQA datasets provide only question/answer text (e.g., Q: what happened after the dog barked? A: he jumped) --- they do not provide temporal spans, or may provide a single, coarse span without explanation, making them insufficient for chain-of-thought evidence grounding.
Temporal grounding datasets contain temporal spans for a query (e.g., Query: ``A barking dog'', Window: ``4-9 seconds''), directly expressing the evidence to localize. However, questions in VideoQA datasets are more complex and do not reveal the evidence needed to determine the answer. We therefore propose a framework to construct evidence chains for videos. See Fig.~\ref{fig:teaser} (right).

Our approach works as follows. First, we generate a pool of potential evidence from segmented video intervals using off-the-shelf VLMs (Sec.~\ref{sec:evidence_pool}). Next, we filter and search through the evidence pool to identify the \emph{most plausible} evidence chain using a combination of LLMs and VLMs (Sec.~\ref{sec:evidence_search}). Finally, we distill this reasoning into our temporally-aware VLM model by training a model to generate the evidence chains (Sec.~\ref{sec:evidence_distill}).

\subsection{Generating the Evidence Pool}
\label{sec:evidence_pool}
We begin by generating a pool of potential evidence $E$ for videos split into multiple segments and across a hierarchy of temporal granularities.
Uniform segmentation or sparse sampling might miss key details because evidence may be unevenly distributed across a video, occurring at different granularities (e.g., static scenes vs. rapid actions). 
To address this, we propose a non-uniform segmentation of the video across $N$ hierarchical levels. In each level, we create a sequence of sub-clips, each of length $L$, and separated by stride $S$ between them.
In our setup, we use $N=5$, $(L,S)\in\{(\sfrac{1}{16}, \sfrac{1}{16}), (\sfrac{1}{8}, \sfrac{1}{16}), (\sfrac{1}{4}, \sfrac{1}{8}), (\sfrac{1}{2}, \sfrac{1}{4}), (1, 1)\}$.\footnote{Both $L$ and $S$ are expressed as fractions of the full video duration.} This approach results in a set of video segments $\{vc_1, \dots, vc_n\}$, covering five levels of granularity from global (full video context with $L=1$ and $S=1$) to fine-grained (small, localized segments with $L=\sfrac{1}{16}$ and $S=\sfrac{1}{16}$).
See Fig.~\ref{fig:evidence_overview} (left) and Appendix~\ref{app:evidence_pool} for details of the hierarchy.

We construct our evidence pool by prompting a VLM (LLaMA-3.2-Vision-Instruct-11B~\cite{dubey2024llama3herdmodels}) to generate evidence for each variable-length segment $vc_i$. In short, the model is asked to ``\textit{describe the contents of this segment that are relevant to the given question (but do not simply answer the question)}.'' The full prompt is in Appendix~\ref{app:prompt_template}. 

Fig.~\ref{fig:evidence_overview} (left) illustrates our evidence pool generation pipeline. After this process, we are left with a pool of candidate evidence $E = \{ev_1, ... , ev_m\}$ from five hierarchical levels that captures various temporal granularities of events in the full video, where each $ev_i = (t_s, t_e, \epsilon)_i$ representing the start time, end time, and evidence text for the visual chunk $vc_i$, respectively.
We include an example of an evidence pool generated across all granularities in Appendix~\ref{app:evidence_pool}.

\subsection{Refining and Searching for Evidence Chains}
\label{sec:evidence_search}
Each evidence piece $ev_{i}\in E$ may provide only partial information needed to answer the question and may lack explicit connections, such as temporal or cause-effect relationships among evidences. To construct a coherent evidence chain from this large, noisy pool, we propose a novel evidence search algorithm based on a text-only LLM (LLaMA-3.1-8B-Instruct~\cite{dubey2024llama3herdmodels}). This algorithm first narrows down the hypothesis space for possible evidence chains, then applies a beam search to identify the strongest chain.

\noindent \paragraph{Evidence Refinement} To reduce noise, we begin by narrowing down the hierarchical evidence pool $E$ to a reduced candidate pool $E^*$ by ranking candidates directly using the LLM. We provide the full evidence pool to the LLM, and prompt it to \textit{``Provide the evidence that will help reach the answer in a step-by-step manner. Limit your evidence chain to at most $K$ steps.''}
This pool consists of \( K \) evidence segments representing a smaller, more manageable set of segments likely to be relevant to the question \( Q \). The top-\( W \) evidence segments that are most likely to decode the correct answer are then selected from this reduced pool. This initial refinement narrows the search space and provides a focused foundation for constructing evidence chains. See Figure~\ref{fig:evidence_overview} (bottom left).

\noindent \paragraph{Evidence Chain Search} Next, we search over sequences of evidence to identify high-likelihood chains, looking for the most coherent answer paths via iterative beam search. 
Starting from a refined initial beam of evidence segments, we initialize a beam with width $W = K / 2$, half the size of the refined evidence pool.
In each iteration, new evidence segments are appended to existing chains within the beam, generating expanded chains that may improve the likelihood of reaching the correct answer. Each expanded chain is then scored with the LLM with its likelihood of supporting the correct answer recalculated.
Chains that meet a specified probability threshold \( T \) are retained as potential candidates, and the beam is updated to only include the top-\( W \) evidence chains based on likelihood scores. The evidence search process is summarized in Algorithm~\ref{alg:evidence_chain_search}. 

This process continues until either an evidence chain exceeds the threshold probability \( T \) or a fixed number of iterations is reached. The algorithm ultimately outputs the evidence chain \( C^* \) with the highest likelihood of correctly answering the question, ensuring a well-supported and coherent response. See Figure~\ref{fig:evidence_overview} (top right).

\begin{algorithm}[t]
\caption{Evidence Chain Search}
\label{alg:evidence_chain_search}
\begin{algorithmic}[1]
\State \textbf{Input:} Question $Q$, Answer $A$, Evidence Pool $E^* = \{ev_1, \dots, ev_m\}$, Beam Width $W$, Threshold $T$
\State \textbf{Output:} Optimal Evidence Chain $C^*$

\State \textbf{Initialize:} Evidence chain $C \gets \emptyset$, Beam $B \gets \{\}$
\State \textbf{Beam Search}
\State Initialize beam $B \gets \{ ev_i : \text{top-}W \text{ by } P(A | Q, ev_i) \}$

\While {any $C_i$ in $B$ is updated}
    \For {each chain $C_i \in B$}
        \State Expand $C_i$ by adding $ev_j \in E^* \setminus C_i$ \label{alg:evidence_chain_search:expand_chain}
        \State Compute $P(A | Q, C_i \oplus ev_j)$ \label{alg:evidence_chain_search:compute_likelihood}
        \State If $P(A | Q, C_i \oplus ev_j) > T$, update $C_i$ \label{alg:evidence_chain_search:update_chain}
    \EndFor
    \State Update $B \gets \{ C_i \in B : \text{top-}W \text{ by } P(A | Q, C_i) \}$
\EndWhile
\State Set $C^* \gets \arg \max_{C_i \in B} P(A | Q, C_i)$ \label{alg:evidence_chain_search:select_best_chain}
\end{algorithmic}
\end{algorithm}
\vspace{-1em}

\noindent \paragraph{Evidence Chain Summarization and Filtering}
While the evidence chains $C^*$ contain rich information related to the question, each piece of evidence was originally generated independently of each other, lacking event sequence information when simply concatenated. To address this, we summarize the entire evidence chain to be sequence-aware with a natural flow using the same text-based LLM, given its inherent ability to do chain-of-thought reasoning.

Specifically, the LLM consolidates the interval $(t_s, t_e)$, objects, events, and question cues in each evidence segment into a logical reasoning path that explicitly references the time intervals, and derives the final answer through step-by-step reasoning. In short, we prompt the LLM to ``\textit{convert this relevant evidence and its temporal span into a chain-of-thought reasoning based on the video.}'' See Figure~\ref{fig:evidence_overview} (bottom left). The detailed prompt is in Appendix~\ref{app:prompt_template}.

To ensure high-quality reasoning paths, we filter chains based on the LLM's ability to reach the correct answer. Specifically, we retain only those chains \( C^*_i \) that allow the LLM to correctly derive the answer \( A \).
Formally, we define a filtering criterion for a chain \( C^*_i \) as:
\begin{equation}
f(C^*_i) = \begin{cases} 
1, & \text{if } \arg\max_{\hat{A}} \log p(\hat{A} | Q, C^*_i) = A, \\
0, & \text{otherwise,}
\end{cases}
\end{equation}
where \( Q \) is the question, \( A \) is the correct answer, and \( \log p(\hat{A} | Q, C^*_i) \) represents the log-likelihood of candidate answer \( \hat{A} \) given \( Q \) and \( C^*_i \). This ensures that only evidence chains that allow the model to logically connect the question, multi-hop evidence, and answer are retained. Successful chains are then added to the training data for the video model. This process yields a final dataset consisting of a video, question, evidence chain-of-thought, and answer.

\subsection{Distilling Evidence Chains Into a Single Model}
\label{sec:evidence_distill}
Finally, we distill the evidence chain information into our \ours~model through curriculum training. 
For this, we add an extra training stage, temporal evidence distillation, to the traditional instruction tuning in VLM models~\cite{ren2024timechat,zhang2024llavavideo}. 

Starting with a base VLM model, we perform instruction tuning (predicting answer tokens) as Stage-1. This enables the model to answer questions but without supporting evidence. Traditional chain-of-thought techniques (e.g., prompting the model to ``think step-by-step'') may be applied with limited success, as shown in our experiments. To improve this, we introduce \emph{evidence distillation} in Stage-2, where the model learns to predict both the evidence chain (Sec.~\ref{sec:evidence_search}) and answer tokens, enabling it to reason across video segments. In both stages, the model is trained using \emph{next token prediction} with cross-entropy loss to maximize the likelihood of generating evidence and/or answer tokens, conditioned on the video and question.

During inference, the model is required to generate the temporal evidence chain followed by the answer or directly answer the question depending on the input prompt.

In our experiments, we build on top of strong backbone VLM models, namely TimeChat~\cite{ren2024timechat} --- a temporally-aware model suited to our fine-grained hierarchical temporal evidence data and LLaVA-Video~\cite{zhang2024llavavideo} --- a recent, SOTA VideoQA model.

\section{Experiments}
\label{sec:experiments}

\begin{figure}[t]
\centering
\includegraphics[width=\columnwidth]{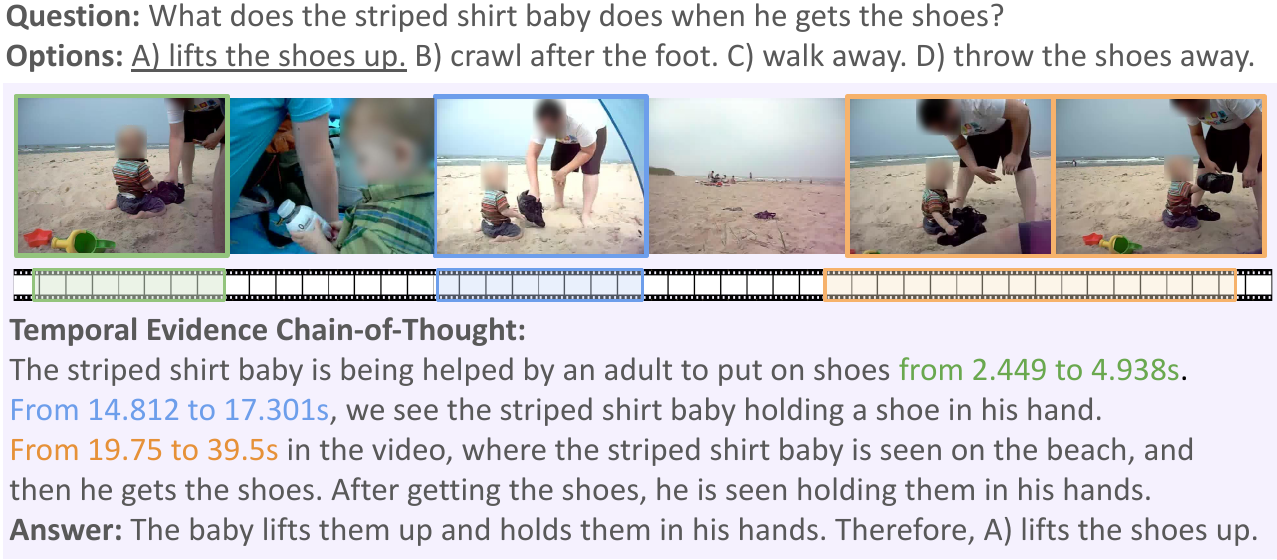}
\caption{Example of temporal evidence on NExT-QA.
}
\label{fig:evidence_example}
\vspace{-4mm}
\end{figure}

\begin{figure}[t]
    \centering
    \begin{minipage}[t]{0.99\columnwidth}
        \centering
        \includegraphics[width=\textwidth]{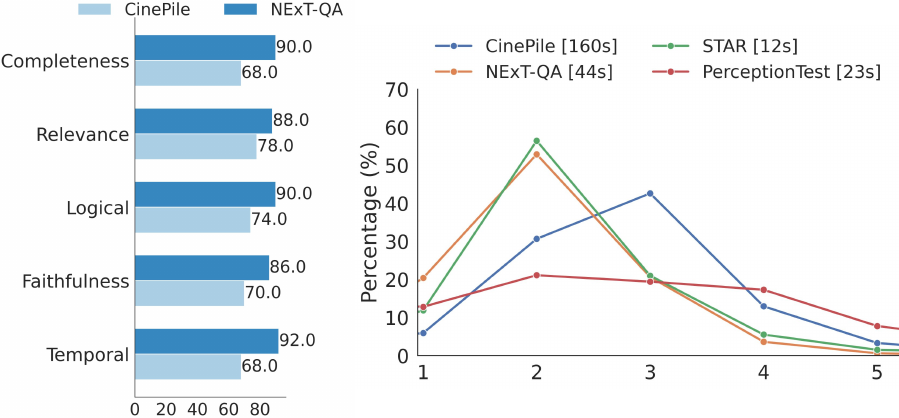}
        \caption{\textbf{Analysis of evidence quality.} \textbf{Left:} Human evaluation score on the quality of temporal evidence chain-of-thought. \textbf{Right:} Distribution of the number of hops in synthesized evidence chain across four datasets.
        }
        \label{fig:evidence_human_eval}
    \end{minipage}
    \hfill
    \vspace{-3mm}
\end{figure}

\begin{table*}[htbp]
    \centering
    \renewcommand{\arraystretch}{1.2}
    \setlength{\tabcolsep}{4pt}
    \resizebox{\linewidth}{!}{
    \begin{tabular}{l>{\kern-0.5\tabcolsep}c cccccccc<{\kern-0.5\tabcolsep}}
    \toprule
    \textbf{Models} & \textbf{Params} & \textbf{CinePile} & \textbf{PercepTest} & \textbf{NExT-QA} & \textbf{STAR} & \textbf{MVBench} & \textbf{NExT-GQA} \\
    \midrule
    \rowcolor{gray!20} \multicolumn{9}{l}{\textit{State-of-the-art Video LLMs}} \\
    SeViLA~\cite{yu2023sevila} & 4B & - & - & 73.8 & 64.9 & - & 16.6 \\
    VideoLLaVA~\cite{lin2024videollava} & 7B & 22.51 & - & - & - & - & - \\
    LongVA~\cite{zhang2024long} & 7B & - & - & 68.3 & - & - & - \\
    LLaMA-3.2V~\cite{dubey2024llama3herdmodels} & 11B & 39.55 & 52.65 & 67.58 & 45.62 & 44.72 & 11.64 \\
    InternVideo2~\cite{wang2024internvideo2} & 6B & - & 63.4 & 78.6 & - & 67.2 & - \\
    LLaVA-OneVision~\cite{li2024llavaonevision} & 7B & 46.42 & 57.1 & 79.4 & 66.24 & 55.91 & 0 \\
    \midrule
    \rowcolor{gray!20} \multicolumn{9}{l}{\textit{TimeChat Base}} \\
    TimeChat~\cite{ren2024timechat} & 7B & 13.67 & 20.34 & 21.71 & 21.03 & 17.08 & 0.00 \\
    TimeChat (Chain-of-Thought) & 7B & 14.27 & 20.96 & 25.60 & 17.54 & 22.05 & 0.00 \\
    TimeChat (Video Instruction Tuning) & 7B & 55.86 & 57.39 & 71.05 & 68.39 & 47.48 & 0.00 \\
    \hdashline
    \ours~ (Dense Caption Distillation) & 7B & 57.85 & 59.94 & 71.94 & 69.99 & 50.12 & 0.00 \\
    \rowcolor{green!5} \ours~ (Temporal Evidence Distillation) & 7B & \bf 58.98 & \bf 63.66 & \bf 73.42 & \bf 70.99 & \bf 50.46 & \bf 27.61 \\
    \midrule
    \rowcolor{gray!20} \multicolumn{9}{l}{\textit{LLaVA-Video Base}} \\
    LLaVA-Video~\cite{zhang2024llavavideo} & 7B & 53.77 & \bf 67.90 & 83.20 & 66.88 & 58.60 & 0.04 \\
    LLaVA-Video (Chain-of-Thought) & 7B & 54.82 & 66.91 & \bf 84.61 & 65.34 & 59.50 & 0.03 \\
    LLaVA-Video (Video Instruction Tuning) & 7B & 54.97 & 67.03 & 83.93 & 67.74 & 59.55 & 0.00 \\
    \hdashline
    \ours~ (Dense Caption Distillation) & 7B & 54.79 & 67.07 & 83.91 & 67.56 & 58.53 & 0.00 \\
    \rowcolor{green!5} \ours~ (Temporal Evidence Distillation) & 7B & \bf 56.39 & 67.50 & 84.13 & \bf 68.23 & \bf 60.95 & \bf 25.19 \\
    \bottomrule
    \end{tabular}
    }
    \caption{\textbf{VideoQA benchmark results.} Our temporal evidence distillation is consistently better than dense caption distillation, direct video instruction tuning, and naive chain-of-thought. \ours~ built on LLaVA-Video achieves SOTA on 4 out of 6 VideoQA benchmarks.
    }
    \vspace{-3mm}
    \label{tab:main_sota}
\end{table*}

\begin{table}[t]
\centering
    \resizebox{\linewidth}{!}{
    \renewcommand{\arraystretch}{1.2}
    \begin{tabular}{>{\kern-0.5\tabcolsep}l ccc<{\kern-0.5\tabcolsep}}
        \toprule
        \multirow{2}{*}{\textbf{Model}} & \multirow{2}{*}{\textbf{LLM}} & \multicolumn{2}{c}{\textbf{NExT-GQA}} \\
        \cmidrule(lr){3-4}
         & & IoP@0.5 & Acc@GQA \\
        \midrule
        LLaMA-3.2V~\cite{li2024llavaonevision} & LLaMA-3~\cite{dubey2024llama3herdmodels} & 19.8 & 11.6 \\
        FrozenBiLM~\cite{yang2022frozenbilm} & DeBERTa~\cite{he2021DeBERTa} & 23.7 & 17.5 \\
        SeViLA~\cite{yu2023sevila} & Flan-T5~\cite{chung2022scalinginstructionfinetunedlanguagemodels} & 22.9 & 16.6 \\
        LLoVi~\cite{zhang2024llovi} & GPT-4~\cite{openai2024gpt4technicalreport} & 38.0 & 26.8 \\
        \midrule
        \rowcolor{green!5} \ours~ & LLaMA-2~\cite{touvron2023llama2openfoundation} & \bf 41.2 & \bf 27.6 \\
        \bottomrule
    \end{tabular}
    }
    \caption{\textbf{Results on NExT-GQA~\cite{xiao2024itrustanswervisually}}. IoP@0.5 and Acc@GQA represent intersection over prediction of evidence and accuracy of grounded question answering.
    }
    \vspace{-3mm}
    \label{tab:ablation_nextgqa}
\end{table}

\begin{table}[t]
\centering
\begin{minipage}{\linewidth}
    \resizebox{\linewidth}{!}{
        \renewcommand{\arraystretch}{1.2}
        \begin{tabular}{>{\kern-0.5\tabcolsep}cc cccc<{\kern-0.5\tabcolsep}}
            \toprule
            \multirow{2}{*}{\bf $\mathbf{E}$-pool} & \multirow{2}{*}{\bf S-CoT} & \multicolumn{4}{c}{\bf NExT-QA}\\
            \cmidrule(lr){3-6}
            & & Temporal & Causal & Descriptive & Avg. \\
            \midrule
             \xmark & \xmark & 68.22 & 71.65 & 74.87 & 71.05 \\
             \cmark & \xmark & 73.50 & 74.35 & 78.87 & 74.79 \\
             \xmark & \cmark & 69.91 & 73.76 & 81.98 & 73.80 \\
            \rowcolor{green!5} \cmark & \cmark & 73.46 & 76.34 & 81.03 & 76.14 \\
            \bottomrule
        \end{tabular}
    }
    \caption{\textbf{Ablation of evidence data} with and without Evidence pool ($E$-pool) or chain-of-thought summarization (S-CoT) stages.}
    \vspace{-3mm}
    \label{tab:tab3_ablation_cot}
\end{minipage}
\end{table}

\begin{table*}[t]
\centering
    \resizebox{\linewidth}{!}{
        \renewcommand{\arraystretch}{1.2}
        \begin{tabular}{>{\kern-0.5\tabcolsep}cc cccccc ccccc cc<{\kern-0.5\tabcolsep}}
            \toprule
            \multirow{2}{*}{\bf Stage-1} & \multirow{2}{*}{\bf Stage-2} & \multicolumn{6}{c}{\bf CinePile} & \multicolumn{5}{c}{\bf STAR} & \multicolumn{2}{c}{\bf NExT-GQA}\\
            \cmidrule(lr){3-8} \cmidrule(lr){9-13} \cmidrule(lr){14-15}
            & & CRD & NPA & STA & TEMP & TH & Avg. & Int. & Seq. & Pre. & Fea. & Avg. & IoP & Acc \\
            \midrule
             \xmark & \xmark & 13.04 & 11.86 & 17.16 & 11.86 & 1.92 & 13.62 & 20.02 & 21.31 & 24.84 & 19.18 & 21.04 & 0.00 & 0.00 \\
             \cmark & \xmark & 57.05 & 54.15 & 58.99 & 44.59 & 63.46 & 55.86 & 63.30 & 70.20 & 74.68 & 72.04 & 68.39 & 0.00 & 0.00 \\
             \xmark & \cmark & 59.59 & 57.20 & 59.81 & 48.30 & 66.99 & 58.12 & 66.93 & 73.16 & 79.03 & 75.92 & 71.76 & 41.60 & 27.42 \\
            \rowcolor{green!5} \cmark & \cmark & 59.80 & 58.89 & 61.57 & 49.23 & 65.38 & 58.98 & 65.76 & 73.06 & 75.80 & 75.31 & 70.99 & 41.22 & 27.61 \\
            \bottomrule
        \end{tabular}
    }
    \caption{\textbf{Stage-1 vs. Stage-2 training.} Effect of Stage-1 (standard video instruction tuning, Q$\rightarrow$A) and Stage-2 (temporal evidence finetuning, Q$\rightarrow$A, Q$\rightarrow$E,A, Q$\rightarrow$A,E) on performance (Sec.~\ref{sec:evidence_distill}).
    }
    \vspace{-3mm}
    \label{tab:tab4_ablation_stage}
\end{table*}

\begin{table}[t]
\centering
    \resizebox{0.98\linewidth}{!}{
    \renewcommand{\arraystretch}{1.15}
    \begin{tabular}{>{\kern-0.5\tabcolsep}l cccc<{\kern-0.5\tabcolsep}}
        \toprule
        \multirow{2}{*}{\textbf{Model}} & \multicolumn{3}{c}{\textbf{NExT-QA}} & \multirow{2}{*}{Avg} \\
        \cmidrule(lr){2-4}
         & Temporal & Causal & Descriptive & \\
        \midrule
        No Evidence & 68.22 & 71.65 & 74.87 & 71.05 \\
        Direct Multi-Evidence & 72.65 & 75.25 & 81.21 & 75.34 \\
        GT-Guided Sampling & 72.36 & 74.31 & 80.50 & 74.64 \\
        \midrule
        \rowcolor{green!5} \ours~        & 73.46 & 76.34 & 81.03 & 76.14 \\
        \quad w/o Hier      & 70.75 & 74.61 & 80.54 & 74.28 \\
        \quad w/o Search    & 69.69 & 73.78 & 78.72 & 73.22 \\
        \quad w/o Multi-Hop & 70.35 & 74.92 & 83.25 & 74.74 \\
        \bottomrule
    \end{tabular}
    }
    \caption{\textbf{Ablation on Evidence Data Framework.} 1) using off-the-shelf model to generate evidence chain, 2) hierarchical evidence pool (Hier), evidence chain search (Search), multi-hop temporal evidence (Multi-Hop).
    }
    \vspace{-4mm}
    \label{tab:tab5_ablation_evidence_pipeline}
\end{table}

We evaluate models on a suite of VideoQA benchmarks, including CinePile~\cite{rawal2024cinepile} for long-video understanding, PerceptionTest~\cite{pătrăucean2023perceptiontest} for low-level video perception, NExT-QA~\cite{xiao2021next}, STAR~\cite{wu2021star_situated_reasoning} and MVBench~\cite{li2024mvbench} for complex reasoning, and NExT-GQA~\cite{xiao2024itrustanswervisually} for temporal evidence grounding. For all datasets, we report MCQ accuracy.

\noindent \paragraph{Baselines}
We compare with state-of-the-art VLMs from existing literature as well as several variants of our approach to show the benefit of our temporal evidence distillation.
\begin{itemize}[leftmargin=*]
\itemsep0em 
\item \textbf{Base VLM} is the off-the-shelf VLM used for our experiments, applied zero-shot. We experiment with \textbf{TimeChat}~\cite{ren2024timechat} and \textbf{LLava-Video}~\cite{zhang2024llavavideo}.
\item \textbf{Chain-of-thought} implements the standard CoT mechanism by adding ``let's think step-by-step'' to the inference prompt. This is to show the non-trivial nature of grounding and generating temporal evidence in current VLMs.
\item \textbf{Video Instruction Finetuning} is the model after standard instruction tuning on VideoQA datasets.
\item \textbf{Dense Caption Distillation} is our model variant trained to generate dense captions (instead of evidence chains) by the same models that generate the evidence pool. This is to assess the value of temporal evidence beyond naive caption augmentation.
\item \textbf{Temporal Evidence Distillation} is our proposed approach in Sec.~\ref{sec:evidence_distill} that trains a model to generate temporal evidence chains alongside direct answers, identifying sequences of temporal spans from multiple relevant parts of the video, together with visual evidence within them.
\end{itemize}
Note that all baselines are trained on the \emph{same data} to allow apples-to-apples comparisons. For completeness, we also compare against state-of-the-art approaches from the literature that benefit from training on larger scale datasets.

\noindent \paragraph{Implementation and Training Details}
We use LLaMA-3.2-Vision-Instruct-11B as the evidence pool generator and LLaMA-3.1-8B-Instruct for evidence refinement and search in Sec.~\ref{sec:evidence_pool} and \ref{sec:evidence_search} respectively. We set beam width $W=4$, threshold $T=0.7$, and the maximum iterations as 3 in Algorithm~\ref{alg:evidence_chain_search}. We use checkpoints from the official code repository as initializations for our TimeChat~\cite{ren2024timechat} and LLaVA-Video~\cite{zhang2024llavavideo} base models.

We train \ours-TimeChat for 10 epochs, and \ours-LLaVA-Video for 1 epoch at each stage.
We include dataset usage at each training stage and evaluation dataset details in Appendix~\ref{app:dataset_detail}. 
\ours~ is trained on 291K video question answering samples from public datasets such as NExT-QA, STAR, and PerceptionTest.
Details of full data mix and additional hyperparameters can be found in Appendix~\ref{app:dataset_detail}.
We use a batch size of 64, with 8 nodes of 8-V100 (32G) machine. For TimeChat-based models, we unfreeze only the image Q-Former, video Q-Former, and linear layer and process 96 input frames for each video, while for LLava-Video, we unfreeze only the adapter and the LLM backbone. We use LoRA~\cite{hu2021lora} with a rank 32 for TimeChat-based and rank 128 for LLaVA-based \ours.

\subsection{Quality of Generated Evidence}
\label{sec:evidence_quality}
To begin, we analyze the quality of evidence chains themselves. First, we verify whether generated evidence chains capture sufficient detail to answer questions (i.e., without re-watching the video). We do this by prompting a text-only LLaMA-3.1-8B-Instruct with a question and an evidence chain, and measuring the accuracy of selecting the correct answer on NeXT-QA.
Through this process, our generated evidence chain yields 72.05\%, significantly improving over the conventional chain-of-thought (51.71\%). 
The remaining gap can be attributed to shortcomings in the base VLM responsible for generating the evidence pool. These issues include hallucinations in video segment descriptions and vague descriptions that omit crucial information, which in turn propagate errors throughout the entire evidence chain.

Next, we evaluate the quality of evidence directly by asking two human annotators to score a subset of evidence chains across five key aspects on a 3-point scale (good, average, bad). These are (1) Temporal: Does the temporal window match the evidence text? (2) Faithfulness: Is the evidence faithful to the video content? (3) Logical: Is the reasoning logical across evidence? (4) Relevance: How relevant is the evidence chain to the video/question? and (5) Completeness: Does the evidence chain capture all required information in the video to answer the question?
Full instructions are in Appendix~\ref{app:human_eval}.
Figure~\ref{fig:evidence_human_eval} (left) presents the average score in each category based on human annotations. We find that while it is harder to generate reliable evidence on datasets with longer videos, our approach still scores over 80.4\% on average across the five aspects, indicating the effectiveness of our synthetic data pipeline. 

Finally, we show statistics of our synthesized evidence data in Figure~\ref{fig:evidence_human_eval} (right). We find that, for benchmarks with longer videos, such as CinePile tend to favor larger number of hops compared with shorter duration video benchmarks, such as STAR and NExT-QA. 
We show a qualitative example of our temporal evidence in Figure~\ref{fig:evidence_example}.
See Appendix~\ref{app:qualitative_eval} for more examples and analysis.

\definecolor{wrong}{rgb}{0.96, 0.79, 0.80}
\definecolor{correct}{rgb}{0.85, 0.91, 0.82}
\definecolor{tempcolor}{rgb}{0.259, 0.522, 0.957}

\begin{figure*}[t]
\centering
\includegraphics[width=\linewidth]{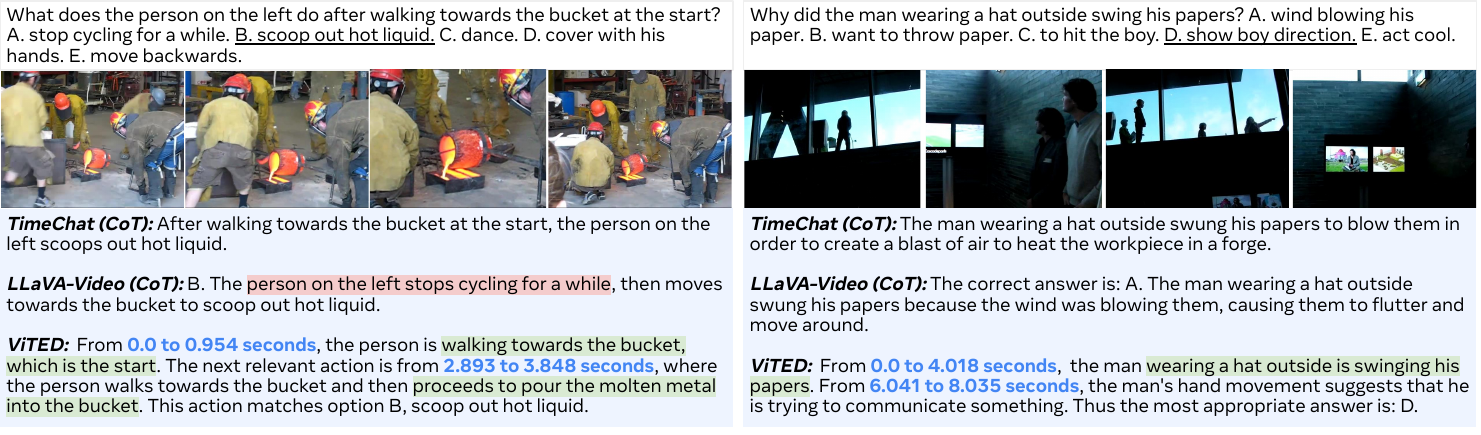}
\caption{\textbf{Examples of generated evidence chains.} Compared to traditional chain-of-thought approaches, \ours~ demonstrates temporal evidence generation and reasoning capabilities, accurately analyzing the sequence of actions in the video to reach the correct final answer. Colored text and highlights are for visualization only and correspond to \hlc[wrong]{wrong evidence}, \hlc[correct]{correct evidence} and temporal localization windows of generated evidence (\textbf{\color{tempcolor}{blue text}}).
}
\label{fig:vqa_model_comparison}
\vspace{-3mm}
\end{figure*}

\subsection{\ours~ for Video Question Answering}
In Table~\ref{tab:main_sota}, our \ours~ with temporal evidence distillation significantly outperforms video instruction tuning across seven video-based QA benchmarks, achieving gains such as $+3.12\%$ on CinePile, $+6.27\%$ on PerceptionTest, and $+27.61\%$ on NExT-GQA with TimeChat-base. Additionally, \ours-LLaVA-Video surpasses the SOTA baseline LLaVA-Video by $+2.62\%$ on CinePile, $+2.35\%$ on STAR, and $+25.19\%$ on NExT-GQA.
Our temporal evidence distillation also consistently surpasses video instruction tuning and dense caption distillation on top of our base video models. This result underscores the model’s strength in handling temporally distributed evidence, which is critical for accurately interpreting video content where events unfold over time.
While LLaVA-Video excels in video question answering and TimeChat in video temporal grounding, neither is tailored for grounding the evidence entailed in questions, leading to near-zero accuracy on NExT-GQA evidence grounding task.

\subsection{\ours~ for Grounding Visual Evidence}
From Table~\ref{tab:main_sota}, \ours~ shows the benefits of temporal evidence capability in achieving strong results on NExT-GQA. This is significant as evidence grounding arises naturally from our model's training, without relying on explicit grounding modules~\cite{chen2024groundedmultihopvideoqalongform,lin2023univtgunifiedvideolanguagetemporal} or manually labeled evidence grounding data~\cite{xiao2024itrustanswervisually,lei-etal-2018-tvqa}. 
We further compare with the current SOTA in evidence grounding on NExT-GQA in Table~\ref{tab:ablation_nextgqa} across two standard metrics, Intersection over Prediction (IoP) and Accuracy (Acc) of grounded question answering (GQA). \ours~achieves a remarkable IoP@0.5 score of 41.22 and Acc@GQA of 27.61\%, outperforming even GPT-4 driven agent approaches like LLoVi~\cite{zhang2024llovi}, and making it the best model to date.

\subsection{Ablation Experiments}
\label{sec:ablations}
We present experiments ablating key design choices in our approach: the importance of different modules in evidence generation (Table~\ref{tab:tab3_ablation_cot}), the role of training stages (Table~\ref{tab:tab4_ablation_stage}), various evidence generation strategies (Table~\ref{tab:tab5_ablation_evidence_pipeline}). Additional ablations on evidence-pool generators and hyper-parameters are in Appendix~\ref{app:quantitative_eval}. Unless otherwise specified, we show single-task fine-tuning results on NExT-QA, using the TimeChat backbone for all the ablations.

\noindent \paragraph{Can we directly generate evidence with VLMs?}
First, we explore simpler methods of evidence generation --- by simply prompting off-the-shelf VLMs to directly generate them. We test two strategies: (1) Direct Multi-Evidence Grounding, where we prompt the model to ``\textit{provide a detailed sequence of information to help answer the question in the form [start time, end time] \{supporting evidence\}}'', given the question and options; and (2) GT-Guided Evidence Sampling, where we sample up to three evidence chains from the same model, and use the ground-truth answer to select the most appropriate one. Full details of each strategy are in Appendix~\ref{app:implementation}. 

Our results in Table~\ref{tab:tab5_ablation_evidence_pipeline} (top) show that both methods are able to produce coherent evidence chains that enhance performance compared to the model with no additional evidence.
Both baseline strategies surpass the baseline without evidence by a large margin, which indicates the importance of evidence. The strategy with the ground truth guidance is even worse, indicating that the input guidance with answer can lead model to make up non-plausible evidence chains.
Meanwhile, we can see our \ours~ surpasses both strategies, indicating that our evidence generation pipeline is effective.
Overall, our approach outperforms these direct variants highlighting the need for more elaborate evidence generation strategies.

In Table~\ref{tab:tab5_ablation_evidence_pipeline} (bottom), we show additional ablations examining the necessity of: hierarchical evidence pools by utilizing only a single level ($S$=1, $L$=1) of the hierarchy (w/o Hier); evidence search by directly using the filtered evidence pool $E^*$ (w/o Search); and multi-hop evidence by forcing evidence to be single hop (w/o MultiHop).
We observe that there is smaller or no performance degradation in ``Descriptive'' type of questions, compared with ``Temporal'' and ``Causal'' types. This indicates the design of our hierarchical evidence pool, evidence search, and multi-hop are essential to complex video understanding.

\noindent \paragraph{Importance of evidence generation stages}
We ablate different stages of the evidence generation pipeline, namely the need for the evidence pool itself (Sec.~\ref{sec:evidence_pool}), the chain-of-thought summarization of evidence (Sec.~\ref{sec:evidence_search}) and their interplay. We simply replace the evidence pool with a single generated evidence chain (similar to the direct approaches above) and/or drop the summarization step from the pipeline, leaving evidence chains in their raw form.

Our results in Table~\ref{tab:tab3_ablation_cot} show that both stages are important to achieve optimal distillation. Although, without the evidence pool, our \ours~ can still achieve on par results as our full model on `Descriptive' category, it is far worse in the `Temporal' and `Causal' category.
While when the summarization module is dropped, the model performance degrades severely in the `Descriptive' category, and slightly on `Causal' category. Without these two evidence generation stages, the model is consistently worse than our full \ours~ across three aspects.

\noindent \paragraph{Are both training Stage-1 and 2 essential?}
In Table~\ref{tab:tab4_ablation_stage}, we ablate Stage-1 and Stage-2 in our proposed temporal evidence curriculum training strategy.
We show accuracy on subcategories: Character and Relationship Dynamics (CRD), Narrative and Plot Analysis (NPA), Setting and Technical Analysis (STA), Temporal (TEMP), Theme Exploration (TH) on CinePile, Interaction (Int.), Sequence (Seq.), Prediction (Pre.), and Feasibility (Fea.) on STAR.
Without Stage-1 and Stage-2 is equivalent to our base TimeChat model. Adding video instruction tuning (Stage-1) leads to significant improvements on CinePile and STAR, but no gain on NExT-GQA which requires evidence grounding. With our Stage-2 (temporal evidence finetuning), we achieve SOTA results on CinePile, STAR and NExT-GQA.

\section{Conclusion}
\label{sec:conclusion}
We proposed a novel pipeline to synthesize high-quality chain-of-evidence data on top of existing video understanding data, and a video model to distill this temporal video evidence data via curriculum training. Our results show notable improvements over state-of-the-art models with larger sizes and more training data, by unlocking temporally-grounded chain-of-thought reasoning in videos.

\section{Acknowledgements}
We sincerely thank Shraman Pramanick for his insightful brainstorming and thoughtful feedback on our early manuscript.

{
    \small
    \bibliographystyle{ieeenat_fullname}
    \bibliography{main}
}

\clearpage
\clearpage
\maketitlesupplementary

\renewcommand{\thesection}{A\arabic{section}}
\renewcommand{\theHsection}{A\arabic{section}}
\renewcommand{\thetable}{A\arabic{table}}
\renewcommand{\thefigure}{A\arabic{figure}}

This section contains supplementary material to support the main paper. The contents include:

\begin{itemize}[leftmargin=*]
\itemsep0em 
    \item (\ref{app:preliminary_study}) Preliminary study on how often evidence chains are actually needed for VideoQA.
    \item (\ref{app:evidence_pool}) Evidence pool generation details and hyper-parameters, and an illustration of our hierarchical evidence pool (Table~\ref{tab:pool_hier12}).
    \item (\ref{app:prompt_template}) Details of full prompts used throughout Sec.~\ref{sec:sec3_approach}.
    \item (\ref{app:implementation}) Baseline details and training setting details to supplement Sec.~\ref{sec:experiments}.
    \item (\ref{app:dataset_detail}) Dataset details including training and evaluation set composition and splits.
    \item (\ref{app:human_eval}) Evaluation protocol details for the human study of evidence quality to supplement Sec.~\ref{sec:experiments}.
    \item (\ref{app:quantitative_eval}) Additional ablation experiments to supplement Sec.~\ref{sec:experiments}.
    \item (\ref{app:qualitative_eval}) Expanded sets of qualitative results to add to those presented already in Figures~\ref{fig:evidence_example} and ~\ref{fig:vqa_model_comparison}.
    \item (\ref{app:discussions}) A discussion on the scope and limitations of our approach.
\end{itemize}

\noindent A supplementary html page with video versions of paper figures is also attached. 

\section{How often are evidence chains required?}
\label{app:preliminary_study}

In Sec.~\ref{sec:sec3_approach} we highlighted the need for evidence chain reasoning.
We investigate what percentage of the questions in current VideoQA datasets actually require evidence chains to reason about the final answer.
We manually annotate $50$ samples and found that $54\%$ (27/50) of videos in NExTQA-val do indeed require evidence. The remaining samples can be answered by single frame or a textual caption from the video itself.
We further investigate how visual evidence is distributed across the video by computing an \emph{entailment score} --- a score for how likely the description of different parts of the video entail the answer of the question. We calculate this by prompting an LLM and validate that high-scoring video segments do contain essential evidence to help question answering.

\section{Evidence Pool Generation Details}
We present additional details and examples of our evidence pool generation strategy to supplement Sec.~\ref{sec:evidence_pool}.

\label{app:evidence_pool}
\subsection{Hyper-parameters}
\label{app:evidence_pool_hyperparam}
Table~\ref{tab:parameters_summary} summarizes the key hyper-parameters across the three main phases: Evidence Pool Generation (Sec.~\ref{sec:evidence_pool}), Evidence Search (Sec.~\ref{sec:evidence_search}), and Model Training (Sec.~\ref{sec:evidence_distill}). 

\paragraph{Evidence Pool Generation} involves hierarchical segmentation with $N=5$ levels ensures evidence extraction across temporal granularities using LLaMA-3.2-Vision-Instruct-11B and a tailored prompt. 

\paragraph{Evidence Search} employs LLaMA-3.1-8B-Instruct with a beam width $W=4$, threshold $T=0.7$, and a maximum of 3 iterations for evidence refinement. 

\paragraph{Model Training} involves two stages: instruction tuning and temporal evidence distillation, with distinct configurations for TimeChat and LLaVA-Video. Key components, including Q-Formers and adapters, are optimized with LoRA ranks of 32 and 128, ensuring reproducibility of the pipeline.

\begin{table*}[t]
\centering
\small
\resizebox{\linewidth}{!}{
\renewcommand{\arraystretch}{1.2}
\begin{tabular}{cc p{11cm}}
    \toprule
    \multicolumn{2}{c}{\bf Parameter} & \multicolumn{1}{c}{\bf Value/Details} \\
    \hline
    \rowcolor{gray!20} \multicolumn{3}{c}{\bf Evidence Pool Generation} \\
    \hline
    \multicolumn{2}{c}{Hierarchical Levels ($N$)} & 5 \\
    \multicolumn{2}{c}{Segment Length ($L$) and Stride ($S$)} & \{(\(\sfrac{1}{16}, \sfrac{1}{16}\)), (\(\sfrac{1}{8}, \sfrac{1}{16}\)), (\(\sfrac{1}{4}, \sfrac{1}{8}\)), (\(\sfrac{1}{2}, \sfrac{1}{4}\)), (1, 1)\} \\
    \multicolumn{2}{c}{Evidence Model} & LLaMA-3.2-Vision-Instruct-11B \\
    \hline
    \rowcolor{gray!20} \multicolumn{3}{c}{\bf Evidence Search} \\
    \hline
    \multicolumn{2}{c}{Refinement Model} & LLaMA-3.1-8B-Instruct \\
    \multicolumn{2}{c}{Beam Width ($W$)} & 4 \\
    \multicolumn{2}{c}{Probability Threshold ($T$)} & 0.7 \\
    \multicolumn{2}{c}{Maximum Iterations} & 3 \\
    \hline
    \rowcolor{gray!20} \multicolumn{3}{c}{\bf Model Training} \\
    \hline
    \multicolumn{2}{c}{Base Models} & TimeChat, LLaVA-Video \\
    \multicolumn{2}{c}{Training Stages} & Stage-1: Instruction Tuning (Answer tokens), Stage-2: Temporal Evidence Distillation (Evidence + Answer tokens) \\
    \multicolumn{2}{c}{Evidence Chain Filtering Criterion} & Correct answer likelihood (\(f(C^*_i) = 1\)) \\
    \multicolumn{2}{c}{Epochs per Stage} & TimeChat: 10 epochs; LLaVA-Video: 1 epoch \\
    \multicolumn{2}{c}{Model Parameters to Train} & TimeChat: Image Q-Former, Video Q-Former, Linear Layer (96 input frames). \newline LLaVA-Video: Adapter, LLM Backbone. \\
    \multicolumn{2}{c}{LoRA Rank} & TimeChat: 32; LLaVA-Video: 128 \\
    \bottomrule
\end{tabular}
}
\caption{\textbf{Summary of Parameters and Settings.} Key parameters used in Evidence Pool Generation, Evidence Search, and Model Training. Metrics are detailed to ensure reproducibility.}
\label{tab:parameters_summary}
\end{table*}

\subsection{Hierarchical Visual Evidence Chunk}
\label{app:evidence_pool_visual_evidence}

The hierarchical narrated evidence approach divides a video into multiple levels of granularity, ranging from a global context to fine-grained, localized details. This structured segmentation allows for capturing evidence across different temporal scales. The five hierarchical levels are as follows:

\begin{itemize}
    \item \textbf{Hier1 (Global)}: Represents the entire video as one single segment, providing a broad context for understanding the overall scene. This level covers the full video duration with $L=1$ and $S=1$, capturing general, high-level information about the content.
    
    \item \textbf{Hier2 (Sectional)}: Divides the video into two large segments, each covering half of the video. This level is used to capture larger shifts or sections of the video, such as changes in scenes, major actions, or transitions. Here, $L=\sfrac{1}{2}$ and $S=\sfrac{1}{4}$.

    \item \textbf{Hier3 (Detailed)}: Breaks the video into smaller chunks (e.g., one-quarter of the video duration), allowing for a more detailed view of specific events and interactions. This level captures important actions or events that may not be apparent at a larger granularity. In this case, $L=\sfrac{1}{4}$ and $S=\sfrac{1}{8}$.
    
    \item \textbf{Hier4 (Fine-Grained)}: Divides the video into even smaller segments, focusing on precise details such as body language, gestures, or smaller interactions. This level allows for a closer examination of fine-grained events. The segment length is $L=\sfrac{1}{8}$ with a stride of $S=\sfrac{1}{16}$.
    
    \item \textbf{Hier5 (Atomic)}: The smallest possible temporal chunk, capturing the finest details of the video. At this level, the model focuses on very localized moments such as rapid actions or fleeting events. Here, $L=\sfrac{1}{16}$ and $S=\sfrac{1}{16}$.
\end{itemize}

This hierarchical segmentation method ensures that evidence from a wide range of temporal scales is captured, from broad contextual understanding to very specific and rapid events.
We showcase one example of our hierarchical evidence pool in Table~\ref{tab:pool_hier12}.

\begin{table*}[h!]
\centering
\small
\resizebox{\linewidth}{!}{
\begin{tabular}{|p{0.15\textwidth}|p{0.15\textwidth}|p{0.15\textwidth}|p{0.15\textwidth}|p{0.15\textwidth}|p{0.15\textwidth}|}
\hline
\multicolumn{1}{|c}{\textbf{Frame 1}} & \multicolumn{1}{|c}{\textbf{Frame 2}} & \multicolumn{1}{|c}{\textbf{Frame 3}} & \multicolumn{1}{|c}{\textbf{Frame 4}} & \multicolumn{1}{|c}{\textbf{Frame 5}} & \multicolumn{1}{|c|}{\textbf{Frame 6}} \\ \hline

\includegraphics[width=0.15\textwidth]{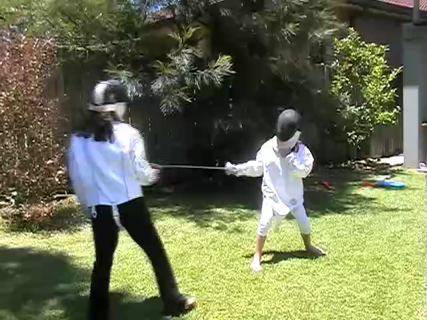} & 
\includegraphics[width=0.15\textwidth]{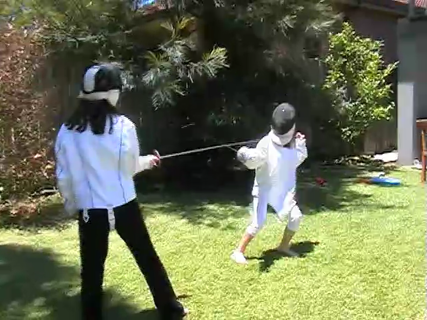} & 
\includegraphics[width=0.15\textwidth]{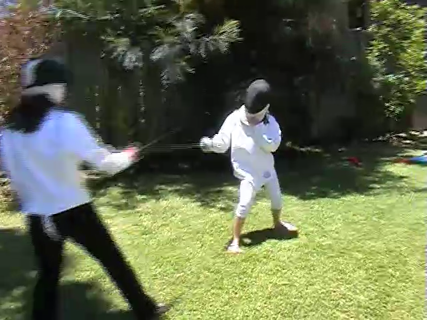} & 
\includegraphics[width=0.15\textwidth]{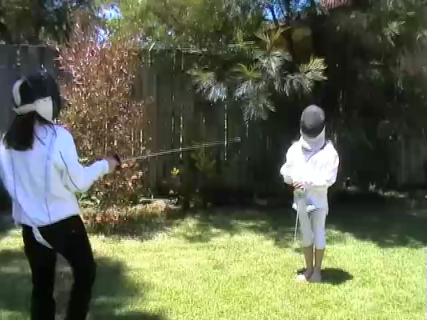} & 
\includegraphics[width=0.15\textwidth]{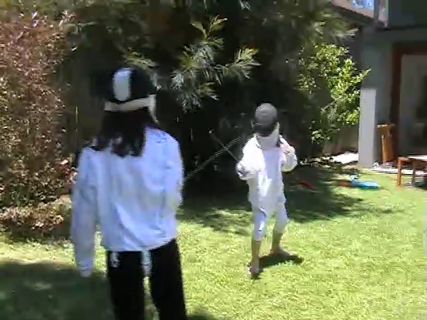} & 
\includegraphics[width=0.15\textwidth]{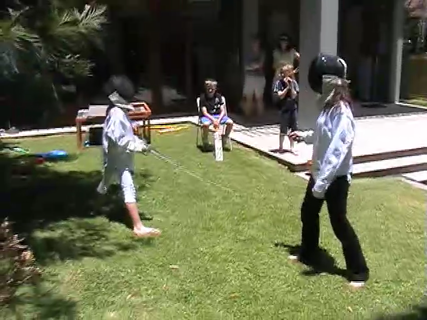} \\ \hline

\multicolumn{6}{|p{\textwidth}|}{
\textbf{Question:} Why are there people surrounding the two people fencing? Options: A. guide. B. performing. C. watching them. D. play for fun. E. to maintain the distance.
} \\ \hline
\rowcolor{gray!20} \multicolumn{6}{|l|} {\textbf{Hierarchy 1:}}\\
\multicolumn{6}{|p{\textwidth}|}{
0. [0.0-0.062seconds] There are no people surrounding the two people fencing. The image only shows two individuals engaging in a fencing match, with no additional individuals present. \newline
1. [0.062-0.125seconds] There are no people surrounding the two people fencing. The image shows a person fencing with another person in a yard, with no other people visible. \newline
... \newline
15. [0.938-1.0seconds] The video shows a scene where two individuals are engaging in a fencing match, with a group of people standing around them. The people surrounding the fencers appear to be watching the match, possibly as spectators or participants in the activity. The presence of these people suggests that the fencing match is part of a larger event or gathering, such as a tournament, training session, or social gathering. The fact that the people are standing around the fencers, rather than actively participating in the match, implies that they are observing the activity rather than engaging in it themselves.
} \\ \hline

\rowcolor{gray!20} \multicolumn{6}{|l|} {\textbf{Hierarchy 2:}}\\
\multicolumn{6}{|p{\textwidth}|}{
0. [0.0-0.125seconds] The video shows two people fencing in a backyard, with a fence surrounding the area. The presence of the fence suggests that the two people are fencing in a designated area, possibly to ensure safety and prevent damage to surrounding objects. \newline
1. [0.062-0.188seconds] The image shows two people wearing fencing masks and holding fencing weapons, with a third person standing nearby. The third person is likely a coach or instructor, providing guidance and support to the two fencers. \newline
... \newline
13. [0.812-0.938seconds] The image depicts two individuals engaged in a fencing match, with a third person observing from a distance. The presence of the third person suggests that they may be a coach, instructor, or spectator, providing guidance and support to the fencers.
} \\ \hline
\rowcolor{gray!20} \multicolumn{6}{|l|} {\textbf{Hierarchy 3:}}\\
\multicolumn{6}{|p{\textwidth}|}{
0. [0.0-0.25seconds] There are no people surrounding the two people fencing. The image shows two people fencing in a backyard, with a house and trees in the background. The two people are wearing fencing gear and holding swords, and appear to be engaged in a fencing match. There is no evidence of people surrounding them. \newline
1. [0.125-0.375seconds] There are no people surrounding the two people fencing. The image only shows two people fencing, and there is no indication of any other individuals present. \newline
... \newline
6. [0.75-1.0seconds] The image depicts two individuals engaging in a fencing match, with a third person standing nearby, likely serving as a referee or instructor. The presence of the third person is likely due to the fact that the two fencers are children, and the third person is there to ensure their safety and provide guidance on proper fencing techniques.
} \\ \hline

\rowcolor{gray!20} \multicolumn{6}{|l|} {\textbf{Hierarchy 4:}}\\
\multicolumn{6}{|p{\textwidth}|}{
0. [0.0-0.5seconds] The video shows two people engaging in a fencing match, with one person holding a foil and the other holding an epee. The scene is set in a backyard, with trees and bushes visible in the background. There are no people surrounding the two fencers. \newline
1. [0.25-0.75seconds] The image shows two people fencing in a backyard, with one person wearing a white shirt and black pants, and the other wearing a white shirt and white pants. The person on the left is holding a fencing sword, while the person on the right is holding a fencing mask. There are no people surrounding them in the image. \newline
2. [0.5-1.0seconds] The individuals surrounding the two people fencing are likely wearing protective gear, including masks, gloves, and fencing jackets, which are characteristic of the sport of fencing. This protective gear is worn to prevent injury during the match. The presence of these individuals, likely coaches or referees, suggests that they are involved in the fencing competition, possibly officiating or providing guidance to the participants.
} \\ \hline

\rowcolor{gray!20} \multicolumn{6}{|l|} {\textbf{Hierarchy 5:}}\\
\multicolumn{6}{|p{\textwidth}|}{
0. [0.0-1.0seconds] The image shows two people fencing, with one person holding a sword and the other person holding a foil. The person on the left is wearing a white shirt and black pants, while the person on the right is wearing a white shirt and white pants. The person on the right is holding a sword in their right hand, while the person on the left is holding a foil in their left hand. \newline
1. [0.0-1.0seconds] There are no people surrounding the two people fencing in the image. The background of the image shows a grassy area with trees and bushes, suggesting that the fencing is taking place outdoors. There is no indication of any people or objects surrounding the two fencers.
} \\ \hline

\end{tabular}
}
\caption{\textbf{Example of our Hierarchical Evidence Pool.} All five levels are shown with generated descriptions. Note: [0.0-1.0 seconds] is normalized to video length.}
\label{tab:pool_hier12}
\end{table*}

\clearpage
\section{Detailed Prompts for Evidence Generation}
\label{app:prompt_template}
We provide detailed prompts corresponding to the high-level prompts in the main paper, specifically, textual evidence narration from video chunks in Sec.~\ref{sec:evidence_pool}, converting evidence to chain-of-thought and evidence candidate refinement in Sec.~\ref{sec:evidence_search}.

\begin{tcolorbox}[colframe=gray,title=Textual evidence narration from video chunks,boxrule=0.5pt]
\small
\textbf{[Video]} Sampled Video Frames \\
\textbf{[Instruction]}
Please provide short and concise evidence from the video that can help answer the question. The format should be as follows:\\
\texttt{Evidence: your\_evidence\_here}\\
\textbf{[Output]} Evidence
\label{box:template}
\end{tcolorbox}
The evidence for each chunk is then paired with its temporal window between $[0,1]$, normalized by the duration of the video (e.g., \texttt{[start-end] evidence}).

\begin{tcolorbox}[colframe=gray,title=Evidence candidate refinement,boxrule=0.5pt]
\small
\textbf{[Video]} Sampled Video Frames

\textbf{[Instruction]}
Use the following video transcript to gather a list of evidence to help answer the question ``{question}". Options: {options}

Transcript:
{transcript}

Provide the evidence in the following json format that will help reach the answer in a step by step manner.
Format:
\begin{verbatim}
{
    "evidence_chain": [
        {
            "start_time": float,
            "end_time": float,
            "evidence": str
        },
        ...
    ]
}
\end{verbatim}

Limit your evidence chain to at most {beam\_width} steps. Respond directly with the json. Please return the evidence as a valid JSON object with proper formatting. Ensure all strings are enclosed in double quotes (") and no invalid syntax is used.

\textbf{[Output]} Evidence
\label{box:template}
\end{tcolorbox}

\begin{tcolorbox}[colframe=gray,title=Converting evidence to chain-of-thought,boxrule=0.5pt]
\small
\textbf{[Video]} Sampled Video Frames

\textbf{[Instruction]}
You're the assistant to seek the visual evidence chain from the video to answer the question ``{question}"

Visual Evidence Observed from Video:
{transcript}

The total duration of the video is {vid\_duration}. Each evidence is the narrated question-relevant information within the [t1-t2seconds] interval of the video.

Please utilize both the timestamps of the evidence and the temporal hint in the question, and also focus on the objects/events in the evidence that strongly indicate the moment described in the question, and then think step-by-step using the most relevant evidence to derive your answer.
Please rewrite relevant evidence and its temporal span into a chain-of-thought reasoning based on the video. Such as, as the question ask about "what does the man do after he enters the room in the end of the video?", we find that both [t1-t2seconds] and [t3-t4seconds] intervals show the man entering the room, since the question is asking end of the video, we look at the latter interval and find that he is picking up a cup after entering the room, thus the answer is xxx.
Please provide your step-by-step reasoning full\_chain\_of\_thought and keep the [t1-t2seconds] when you describe the visual evidence. You can merge [t1-t2seconds] and [t3-t4seconds] as [t1-t4seconds] when they're the same evidence information. Based on your step-by-step reasoning, select the most appropriate option letter as your final\_answer. Please try to only include the evidence that is relevant and necessary for answering the question.
Format:
\begin{verbatim}
{
    "full_chain_of_thought": str,
    "final_answer": str
}
\end{verbatim}
Respond directly with the JSON.\\

\textbf{[Output]} Evidence
\label{box:template}
\end{tcolorbox}

\section{Additional Implementation Details}
\label{app:implementation}

We provide more details for two of our baselines in Sec.~\ref{sec:experiments} below.
\paragraph{Direct Multi-Evidence Grounding} This method tests whether LLaMA-3.2-Vision-Instruct-11B can replace a hierarchical evidence pool and multi-hop search by directly generating temporally-aware, multi-hop evidence in one pass. We prompt the model to output evidence by referencing specific video timestamps, formatted as [start\_time-end\_time], with accompanying descriptions. This approach assesses if the model can synthesize detailed, multi-hop evidence without the need for additional structuring or search processes.
    Prompt: ``\texttt{[Question]} \texttt{[Option]} Please provide detail sequence of information of each part of the video that help answering the question. The format should be in the form of: [start\_time2-end\_time2] This clip 1 shows that xxx which indicate xxx. [start\_time2-end\_time2] This clip 2 shows that xxx which indicate xxx...: …''
\paragraph{GT-Guided Evidence Sampling}
    This strategy uses GT-based filtering to iteratively refine evidence. The model initially generates an evidence chain in response to a question and potential answers, then reviews the evidence against the video data to identify the answer most aligned with the ground truth. If the generated evidence chain does not yield the correct answer, we adjust the model’s response by varying temperature settings or re-prompting, allowing up to three iterations to improve the evidence quality. This method evaluates whether the model can consistently generate valid, temporally-grounded evidence chains using direct GT guidance.
    Prompt: ``\texttt{[Question]} \texttt{[Option]} Please provide your evidence chain in order in the video that help answering the question.''

\section{Training and Evaluation Dataset Details}
\label{app:dataset_detail}
We provide more details about the training and evaluation datasets used in Sec.~\ref{sec:experiments}.

\paragraph{Training Data}
For Stage-1 training, we use the training splits of PerceptionTest (7.4K), NExT-QA (34.1K), STAR (45.7K) and a collection of 127.1K instances from public long video QA datasets.
For Stage-2, we train with additional instances of our temporal evidence data synthesized by LLaMA-3.2-Vision-Instruct-11B and LLaMA-3.1-8B-Instruct following Sec.~\ref{sec:evidence_pool} and \ref{sec:evidence_search}, bringing the total to 291K. Specifically, PerceptionTest (12.3K), NExT-QA (58.3K), STAR (75.7K) and long video QA (145.1K).

\paragraph{Evaluation Data}
We evaluate our models on the validation splits of CinePile~\cite{rawal2024cinepile}, PerceptionTest~\cite{pătrăucean2023perceptiontest}, NExT-QA~\cite{xiao2021next}, STAR~\cite{wu2021star_situated_reasoning}, MVBench~\cite{li2024mvbench}, NExT-GQA~\cite{xiao2024itrustanswervisually}.

\section{Evidence Quality Evaluation Protocol}
\label{app:human_eval}
As mentioned in Sec.~\ref{sec:evidence_quality}, we collect human annotations to verify the quality of evidence chains. We provide detailed instructions and evaluation protocol for the study.
We recruit two graduate students with expertise in video understanding, and ask each of them annotate $50$ examples across five aspects. The instructions for the annotation task are provided below.

\begin{table*}
  \centering
  \resizebox{\linewidth}{!}{
  \begin{tabular}{|p{0.01\linewidth}|p{0.2\linewidth}|p{0.2\linewidth}|p{0.2\linewidth}|p{0.22\linewidth}|p{0.22\linewidth}|}
    \hline
    \textbf{} & \textbf{Temporal} & \textbf{Faithfulness} & \textbf{Logical} & \textbf{Relevance} & \textbf{Completeness} \\
    \hline \rowcolor{green!5}
     3 & The evidence correctly identifies the time sequence of events. & The evidence is fully consistent with the video content. & The evidence forms a coherent and logical reasoning chain. & The evidence is directly relevant to the question and frames. & The evidence includes all critical information needed to answer the question. \\
    \cline{2-6}
    \hline \rowcolor{orange!5}
     2 & The temporal sequence is somewhat accurate but contains minor errors. & The evidence is mostly accurate but includes minor inconsistencies. & The reasoning is partially logical but has gaps or weak links. & The evidence is somewhat relevant but includes unnecessary information. & The evidence captures most key details but omits some minor elements. \\
    \cline{2-6}
    \hline \rowcolor{red!5}
     1 & The evidence significantly misrepresents the time sequence. & The evidence is misleading or contains major inaccuracies. & The reasoning is illogical or lacks coherence. & The evidence is irrelevant or off-topic. & The evidence is incomplete and misses significant details. \\
    \hline
  \end{tabular}
  }
  \caption{\textbf{Scoring rubric for evidence quality.} Annotators score each evidence chain on a three-point scale, across five aspects.}
  \label{app:tab:scoring_rubric}
\end{table*}

\begin{tcolorbox}[colframe=gray,title=Evidence quality annotation instructions,boxrule=0.5pt]
\small
\textbf{Objective:} The purpose of this evaluation is to assess the quality of evidence chains generated for answering video-based questions. Your task is to review the provided evidence chains in the context of the video content and score their quality across five distinct aspects. Your feedback will help refine and improve the performance of the evidence generation system.

\textbf{Workflow:} You will be shown a video and a question related to the video. A generated evidence chain, which includes textual descriptions of events and reasoning, will be presented.
Carefully review the video and identify the key events occurring between them. Read the question and ensure you understand what is being asked. Compare the evidence chain to the video frames and question, evaluating it based on the five aspects described below.

\begin{itemize}[leftmargin=*]
\itemsep0em 
  \item \textbf{Temporal:} Does the temporal window match the evidence text?
  \item \textbf{Faithfulness:} Is the evidence faithful to the video content?
  \item \textbf{Logical:} Is the reasoning logical across the evidence?
  \item \textbf{Relevance:} How relevant is the evidence chain to the video/question?
  \item \textbf{Completeness:} Does the evidence chain capture all required information in the video to answer the question?
\end{itemize}

\noindent \textbf{Important Notes:}
\begin{itemize}[leftmargin=*]
\itemsep0em 
    \item Be as objective as possible. Focus on the content and structure of the evidence chain.
    \item If you are unsure about any aspect, provide comments for clarification.
    \item Your evaluation will directly impact the improvement of the evidence generation system.
\end{itemize}

\label{box:instructions}
\end{tcolorbox}

The scoring rubric is presented in Table~\ref{app:tab:scoring_rubric} and Table~\ref{app:tbl:human_study_examples} shows some examples of the collected human annotations.

\definecolor{tempcolor}{rgb}{0.259, 0.522, 0.957}

\begin{table*}[h!]
\centering
\resizebox{\linewidth}{!}{
\begin{tabular}{|p{0.25\textwidth}|p{0.25\textwidth}|p{0.25\textwidth}|p{0.25\textwidth}|}
\hline
\multicolumn{1}{|c}{\textbf{Frame 1}} & \multicolumn{1}{|c}{\textbf{Frame 2}} & \multicolumn{1}{|c}{\textbf{Frame 3}} & \multicolumn{1}{|c|}{\textbf{Frame 4}} \\ \hline

\multicolumn{1}{|c}{\includegraphics[height=0.15\textwidth]{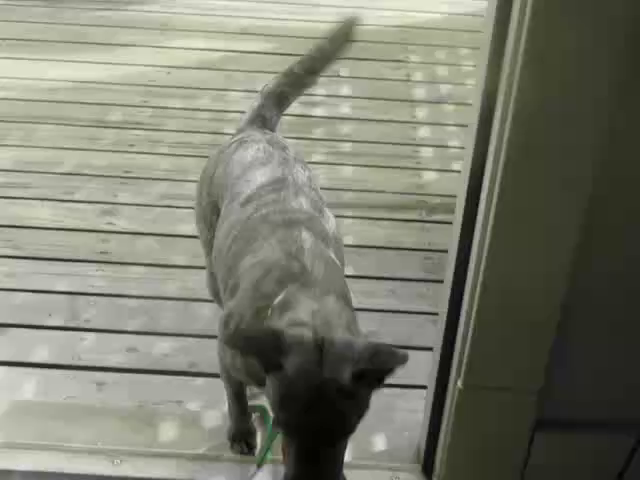}} &
\multicolumn{1}{|c}{\includegraphics[height=0.15\textwidth]{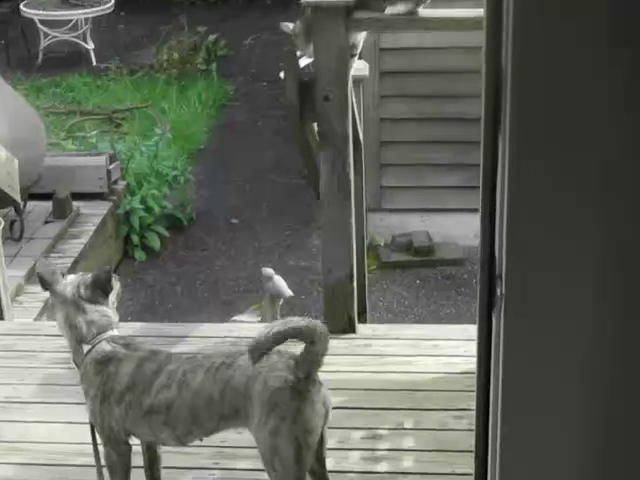}} &
\multicolumn{1}{|c}{\includegraphics[height=0.15\textwidth]{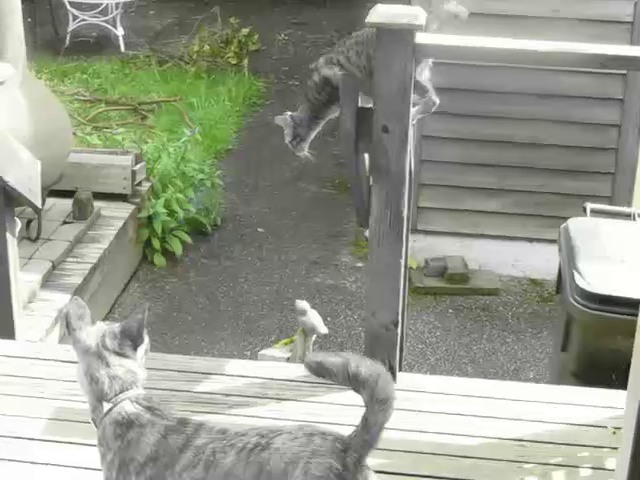}} &
\multicolumn{1}{|c|}{\includegraphics[height=0.15\textwidth]{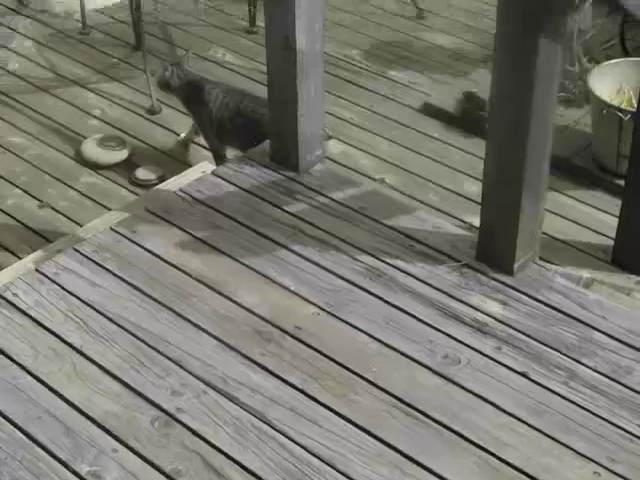}} \\ \hline

\multicolumn{4}{|p{1.1\textwidth}|}{
\textbf{Question:} How many animals are involved in the video? Options: A. five. B. four. C. one. D. three. E. two. \newline
\textbf{Evidence:} As the video starts, we look at the interval from {\textbf{\color{tempcolor}{0.0 to 11.252 seconds}}} in the video, and observe that there is only one animal, a dog. However, this observation does not necessarily indicate the number of animals in the video throughout its duration. We then examine the subsequent intervals to determine the number of animals involved. In the interval from {\textbf{\color{tempcolor}{5.581 to 16.924 seconds}}} in the video, we see a dog and a cat, which suggests that there are at least two animals in the video. Furthermore, in the intervals from {\textbf{\color{tempcolor}{16.924 to 22.505 seconds}}} in the video, and from {\textbf{\color{tempcolor}{22.505 to 28.086 seconds}}}  in the video, we continue to observe a dog and a cat, confirming that there are indeed two animals present in the video. \newline
\textbf{Answer:} Therefore, based on the visual evidence, it is evident that the number of animals involved in the video is two. \newline
\textbf{Scores:} Temporal: 3, Faithfulness: 3, Logical: 3, Relevance: 3, Completeness: 2
} \\ \hline

\multicolumn{1}{|c}{\textbf{Frame 1}} & \multicolumn{1}{|c}{\textbf{Frame 2}} & \multicolumn{1}{|c}{\textbf{Frame 3}} & \multicolumn{1}{|c|}{\textbf{Frame 4}} \\
\hline
\multicolumn{1}{|c}{\includegraphics[height=0.15\textwidth]{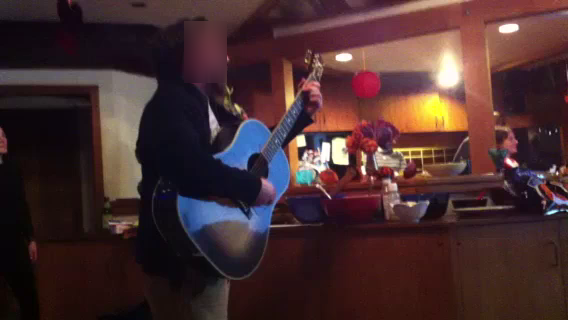}} &
\multicolumn{1}{|c}{\includegraphics[height=0.15\textwidth]{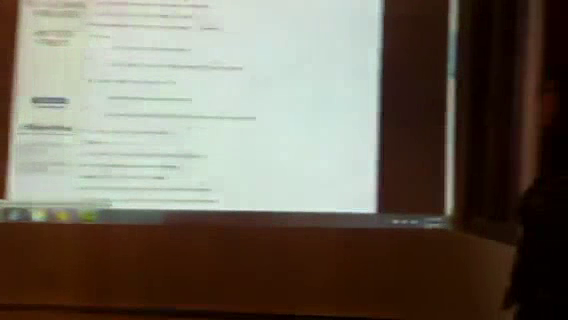}} &
\multicolumn{1}{|c}{\includegraphics[height=0.15\textwidth]{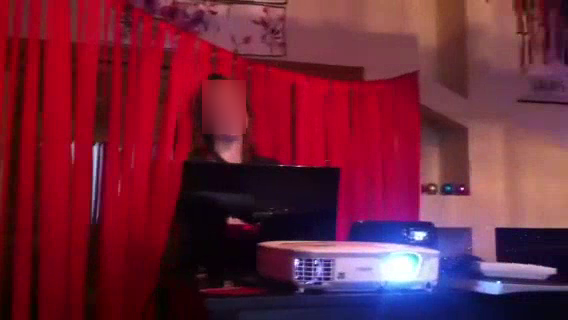}} &
\multicolumn{1}{|c|}{\includegraphics[height=0.15\textwidth]{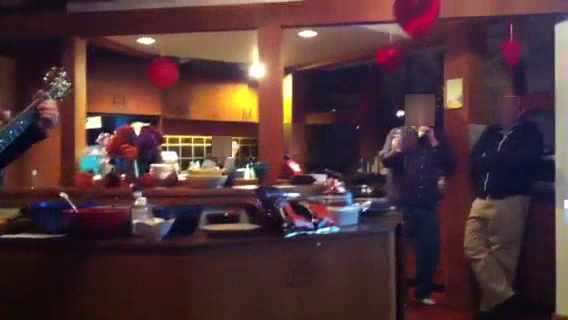}} \\ \hline

\multicolumn{4}{|p{1.1\textwidth}|}{
\textbf{Question:} Why is there a projector on the table? Options: A. displaying lyrics of the song. B. listening to speaker. C. to feed baby. D. eat. E. drink waters. \newline
\textbf{Evidence:} From the given visual evidence, we can observe that the projector on the table is being used for presentation or display purposes from {\textbf{\color{tempcolor}{0.0 to 4.621 seconds}}} and {\textbf{\color{tempcolor}{9.242 to 11.535 seconds}}}. It is also mentioned that the projector is connected to a laptop and is being used to display a presentation or slideshow from {\textbf{\color{tempcolor}{0.0 to 18.485 seconds}}}. However, from {\textbf{\color{tempcolor}{18.485 to 36.97 seconds}}}, the image depicts a man playing a guitar on stage, with a projector positioned on the table in front of him. The projector is likely being used to display visual elements or lyrics for the performance, enhancing the overall experience for the audience. \newline
\textbf{Answer:} This suggests that the projector is being used to display lyrics of the song during the performance. \newline
\textbf{Scores:} Temporal: 3, Faithfulness: 2, Logical: 3, Relevance: 2, Completeness: 3
} \\ \hline

\end{tabular}
}
\caption{\textbf{Human Evaluation of Evidence Quality.} Two examples are provided above. Temporal evidence evidence segments are highlighted for visualization only.}
\label{app:tbl:human_study_examples}
\end{table*}

\section{Additional Ablation Experiments}
\label{app:quantitative_eval}

We present additional ablation experiments to supplement Sec.~\ref{sec:ablations}.

\noindent \paragraph{Evidence Generator and CoT Narrator}
In Table~\ref{tab:tab6_ablation_datagenerator}, we test various evidence generator models and the Chain-of-Thought (CoT) narrator models. Larger models like LLaVA-NEXT-Video-32B achieved the highest average performance, particularly excelling in causal and descriptive questions. While models with a smaller CoT narrator generally performed worse (3B versus 8B in full model), indicating a strong L-CoT model is useful for complex multi-step tasks.

\begin{table}[t]
\centering
\begin{minipage}{\linewidth}
\centering
    \resizebox{0.98\linewidth}{!}{
    \renewcommand{\arraystretch}{1.2}
    \begin{tabular}{>{\kern-0.5\tabcolsep}l cccc<{\kern-0.5\tabcolsep}}
        \toprule
        \multirow{2}{*}{\textbf{Model}} & \multicolumn{3}{c}{\textbf{NExT-QA}} & \multirow{2}{*}{Avg} \\
        \cmidrule(lr){2-4}
         & Temporal & Causal & Descriptive & \\
        \midrule
        \multicolumn{4}{l}{\textbf{Evidence Generator}} \\
        LLaVA-OneVision-0.5B~\cite{li2024llavaonevision} & 72.27 & 75.17 & 81.47 & 75.22 \\
        \rowcolor{green!5} LLaMA-3.2-Vision-Instruct-11B~\cite{li2024llavaonevision} & 73.46 & 76.34 & 81.03 & 76.14 \\
        LLaVA-OneVision-7B~\cite{li2024llavaonevision} & 74.26 & 76.91 & 85.33 & 77.36 \\
        LLaVA-NEXT-Video-32B~\cite{zhang2024llavanext-video} & 74.38 & 78.52 & 83.01 & 77.88 \\
        \midrule
        \multicolumn{4}{l}{\textbf{CoT Narrator}} \\
        LLaMA-3.2-Instruct-3B~\cite{dubey2024llama3herdmodels} & 70.78 & 74.68 & 81.72 & 74.52 \\
        \bottomrule
    \end{tabular}
    }
    \caption{\textbf{Ablation on Evidence Generator} 1) Evidence Generator: model for evidence pool generation, and 2) CoT Narrator: LLM for evidence refinement and searching.}
    \label{tab:tab6_ablation_datagenerator}
\end{minipage}
\end{table}

\section{Additional Qualitative Evaluations}
\label{app:qualitative_eval}
\subsection{Evidence Examples}
\label{app:evidence_examples}

\begin{figure*}[!t]
\centering
\includegraphics[width=\linewidth]{app_assets/vited_evidencedata1.pdf}
\includegraphics[width=\linewidth]{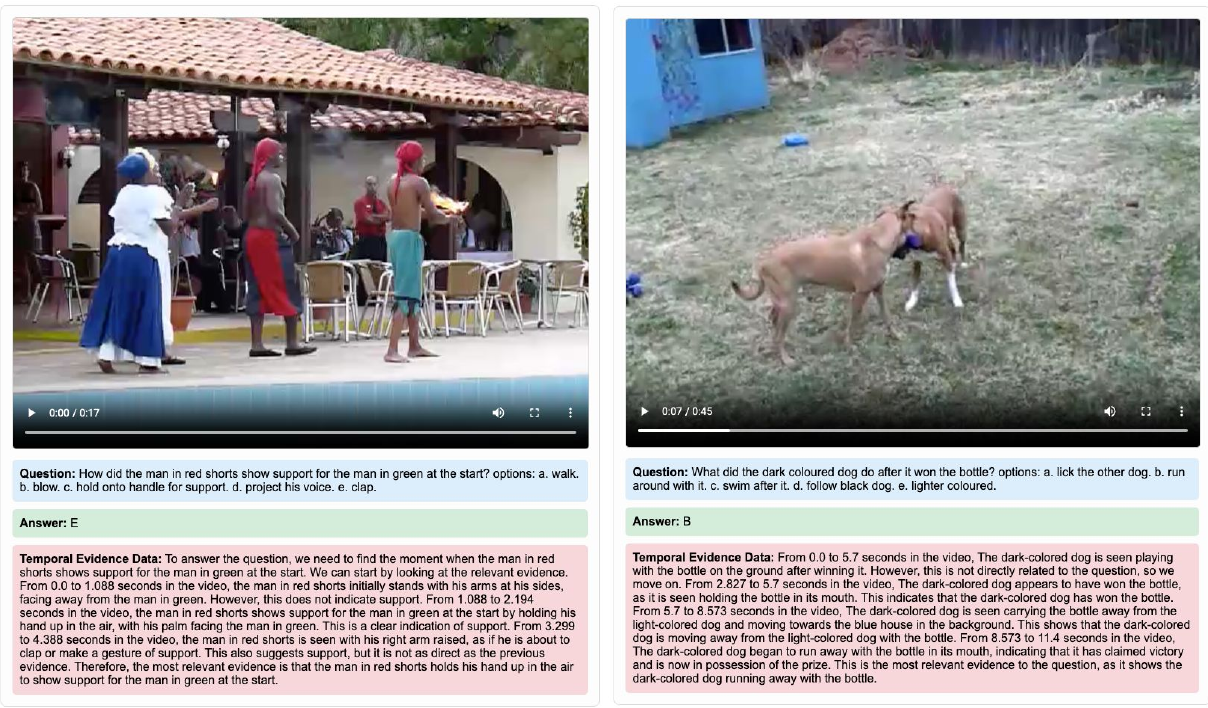}
\caption{Evidence Examples
}
\label{fig:more_evidence_examples}
\end{figure*}

Figure~\ref{fig:more_evidence_examples} illustrates examples of evidence chains generated for video question answering. For more examples, please refer to our `vited\_evidence\_data.html'.

\subsection{Video Question Answering}
\label{app:video_qa}

Figure~\ref{fig:more_vqa_showcases} demonstrates the effectiveness of our approach in answering complex video-based questions. For more examples, please refer to our `vited\_prediction.html'.

\section{Scope and Limitations of \ours}
\label{app:discussions}

\ours~represents a significant step forward in enabling evidence-based temporal reasoning for complex video question answering. However, its scope is inherently influenced by the nature of current video datasets and limitations of large language models (LLMs). The approach excels at scenarios with relatively simple question structures requiring 1-3 reasoning hops, as observed in benchmarks like NExT-QA and STAR. These tasks align well with the hierarchical evidence pool and chain-of-thought generation strategies utilized. However, questions demanding deeper reasoning or involving rare, highly nuanced interactions remain challenging, given the constrained model's ability to synthesize and interpret uncommon event sequences.

Additionally, the reliance on automated evidence generation introduces imperfections. Evidence chains, while effective in many cases, may propagate errors from initial noisy predictions, such as hallucinated or vague descriptions of video segments. These issues can compound, especially in long videos with sparse critical evidence. Furthermore, while the hierarchical evidence framework attempts to capture granular and global contexts, it is not infallible in identifying or relating temporally distant yet causally connected events. Future work could enhance the model's robustness by refining temporal evidence representation and incorporating richer, multi-modal cues to address these limitations.

\begin{figure*}[!t]
\centering
\includegraphics[width=\linewidth]{app_assets/vited_pred1.pdf}
\includegraphics[width=\linewidth]{app_assets/vited_pred2.pdf}
\caption{Video Question Answering Showcases
}
\label{fig:more_vqa_showcases}
\end{figure*}

\end{document}